\documentclass[runningheads]{llncs}
\usepackage{graphicx}
\usepackage{algorithm}
\usepackage{algorithmic}
\usepackage{amsmath}
\usepackage{xspace}
\usepackage{newfloat}
\usepackage{listings}
\newcommand{\oursystem}{\textit{sparql-qa}\xspace}
\begin{document}
\title{Reducing the impact of out of vocabulary words in the translation of natural language questions into SPARQL queries \thanks{This work constitutes a draft pending submission to a journal.}}
\titlerunning{Reducing the impact of WOOV in SPARQL QA}
%
\author{Manuel Borroto\inst{1} \and
Francesco Ricca\inst{1}\orcidID{0000-0001-8218-3178} \and
Bernardo Cuteri\inst{1}\orcidID{0000-0001-5164-9123}}
\authorrunning{M. Borroto et al.}
%
\institute{University of Calabria, Rende (CS) 87036, Italy
\email{\{manuel.borroto,francesco.ricca,bernardo.cuteri\}@unical.it}}
\maketitle              
\begin{abstract}
Accessing the large volumes of information available in public knowledge bases might be complicated for those users unfamiliar with the SPARQL query language. 
Automatic translation of questions posed in natural language in SPARQL has the potential of overcoming this problem.
Existing systems based on neural-machine translation are very effective but easily fail in recognizing words that are Out Of the Vocabulary (OOV) of the training set. 
This is a serious issue while querying large ontologies.
In this paper, we combine Named Entity Linking, Named Entity Recognition, and Neural Machine Translation to perform automatic translation of natural language questions into SPARQL queries. 
We demonstrate empirically that our approach is more effective and resilient to OOV words than existing approaches by running the experiments on Monument, QALD-9, and LC-QuAD v1, which are well-known datasets for Question Answering over DBpedia.

\end{abstract}

\section{Introduction}
Two of the hottest interrelated topics in computer science are linked data and Artificial Intelligence (AI). 
Today we live in the Digital Age, where knowledge is generated and shared at high speed and volume.
As evidence of this, we now have vast and complex knowledge bases that allow gathering large volumes of information through the intercommunication of thousands of datasets referring to various domains in what is known as Linked Data. 
The existence of such knowledge bases means that people have access to a large amount of information never thought, and the DBpedia \cite{lehmann2015dbpedia} project is a real example of that, which is one of the most popular knowledge bases nowadays.
However, the search and retrieval of the information stored in this way can be a hard task for lay users because it is necessary to know the structure of the knowledge base and the appropriate query languages, such as SPARQL \cite{standarsw3c_sparql}.
As a result, AI techniques for natural language Question Answering (QA) have taken a central role in the area of the Semantic Web to address such issues.
Indeed, systems able to translate questions posed in natural language in SPARQL queries have the potential of overcoming this problem because they can remove all technical complexity to the final users.

In recent years, many AI systems have been proposed that are able to translate automatically questions posed in natural language in SPARQL queries~\cite{chen2020formal,panchbhai2020exploring,soru2017sparql}.
The most recent and effective proposals are mostly based on deep neural networks to tackle the problem and exploit the great development achieved by Deep Learning in the last few years.
Indeed, existing systems based on neural-machine translation are very effective but are sensitive to the problem of recognizing words that are Out Of the Vocabulary (OOV) of the training set. Note that OOV becomes a serious problem while querying large ontologies with very many individuals that are not (or cannot be) mentioned in the training phase.

In this paper, we propose an AI system for QA that takes into account more explicitly the problem of dealing with  OOV words.
The core of our architecture is based on a Neural Machine Translation (NMT)~\cite{bahdanau2014neural} module, which is based on Bidirectional Recurrent Neural Networks~\cite{sutskever2014sequence}, \textit{trained side-by-side} with a Named Entity Recognition (NER) module, implementing a BiLSTM-CRF network~\cite{huang2015bidirectional}.
The NMT module translates the input NL question into a SPARQL template, whereas the NER module extracts the entities from the question. The combination of the results of the two modules results in a SPARQL query ready to be executed.
We also introduce a formal definition of a training set format that reduces the output space, and is essential for the proper functioning of the system, and allows us to tackle the problem with OOV words, a major weakness of the majority of the related approaches today.
We empirically test the system on the data sets for question answering on the DBpedia ontology, namely the Monument dataset~\cite{hartmannMarxSoru2018}, QALD-9~\cite{ngomo20189th}, and LC-QuAD v1~\cite{trivedi2017lc}.

\section{Preliminaries}
\subsection{Knowledge Bases and SPARQL}
Informally, a knowledge base is a formal description of a domain of interest that is suitable to be managed by an engine reasoning about the facts modeled in the knowledge base itself, e.g., query existing knowledge or obtain new knowledge.
A formal description of knowledge as a set of concepts within a domain and the relationships that hold between them is called \textit{ontology}~\cite{gruber1995toward}.
Ontologies play a crucial role in computer science, making information interpretable by both humans and machines.
Ontologies are specified by formal languages modeling individuals, classes, attributes, and relations as well as restrictions, rules, and axioms.
An ontology together with a set of individual instances of classes constitutes a \textit{knowledge base} (\textbf{KB}).

Nowadays, there are well-defined languages and technologies to express knowledge bases, de facto standards, such as RDF, RDFS, OWL, are created and supported by W3C~\cite{standarsw3c}.
In RDF, information is usually represented by a collection of \emph{subject-predicate-object} triples $\langle s,p,o \rangle$ where the $p$ establishes a binary relationship between $s$ and $o$. Then a knowledge base is composed of a set of triples, known as RDF-graph \cite{standarsw3c_rdf}.
In a KB, resources can be defined through URIs, which allow reference to non-local resources, enabling the interaction among multiple KBs.

To query RDF-graphs, the standard is SPARQL, an SQL-like language. The syntax and semantics of the language allow the user to query a KB by defining triples looking for a match with \emph{subject-predicate-object} patterns within the graph. The following is an example of a SPARQL query:

{\small
\begin{verbatim}
PREFIX dbo: <http://dbpedia.org/ontology/>
PREFIX dbr: <http://dbpedia.org/resource/>
SELECT ?place WHERE { dbr:Hillary_Clinton dbo:birthPlace ?place }
\end{verbatim}
}

The first and second lines (PREFIX) define the prefix namespace, which is used for abbreviating URIs. The third line (SELECT) returns the values of the variable \emph{?place}. The \emph{WHERE} clause contains the match against the RDF-graph, where \textit{dbr:Hillary\_Clinton} identifies the KB resource, and the identifiers that begin with ``?'' are considered variables.
This query models the answer to the question ``Where was Hillary Clinton born?''.
Hereafter, the answers to a SPARQL query $Q$ are denoted by $answers(Q)$.
The full language allows one to express complex queries. 

\subsection{Bidirectional Recurrent Neural Networks}
In this paper, we rely on the use of Artificial Neural Networks (ANN)~\cite{francois2017deep}, especially \textit{Bidirectional Recurrent Neural Networks} (BRNN)~\cite{schuster1997bidirectional}. 
A particular kind of neural network called \textit{Recurrent Neural Networks} (RNN) has shown a good performance while addressing NLP complex tasks~\cite{giles1994dynamic}. An RNN can operate on variable lengths sequences $S=(x_1, . . . , x_t)$, obtaining an output $y$ that can be of variable size.
In this way, it is possible to process sentences in natural language, considering that they are semantically ordered sequences of words.
An RNN works by iterating over the elements of the sequence $S$ and keeping a state $h$ that contains information relative to what was already processed so that the result of processing the element at time $t$ is also conditioned by the previous information $t-1$ \cite{francois2017deep}. At each time step $t$, the state $h_{(t)}$ is updated by mean of $h_{(t)} = f(h_{(t-1)}, x_t)$, where $f$ is a non-linear function.
This way of operation is useful to capture semantic relationships between the words in a sentence. 

Since it is common to deal with large sequences, there is a phenomenon known as Vanishing Gradient Descent \cite{hochreiter1998recurrent}\cite{bengio1994learning} that affects the performance of RNNs. To address this problem, Hochreiter et. al \cite{hochreiter1997long} proposed a new type of RNN called \textit{Long short-term memory} (LSTM).
LSTM adds a way of transporting information through many time steps. Imagine a conveyor belt running parallel to the sequence you are processing. Information from the sequence can jump onto the conveyor belt at any point, be transported to a later timestep, and jump off, intact, when you need it. This is essentially what LSTM does: it saves information for later, thus preventing older signals from gradually vanishing during processing \cite{francois2017deep}. To this end, LSTMs employ a mechanism called $Gates$ when computing the hidden states. The gating can regulate the flow of information and decide what information is important to keep or throw away. The data processing is done by: 
\begin{align}
\label{eqn:lstm} \footnotesize
   i &= \sigma (x_tU^i + s_{t-1}W^i)\nonumber &
   f &= \sigma (x_tU^f + s_{t-1}W^f)\nonumber\\
   o &= \sigma (x_tU^o + s_{t-1}W^o)\nonumber &
   g &= \tanh (x_tU^g + s_{t-1}W^g)\nonumber\\
   c_t &= c_{t-1}\circ f + g \circ i\nonumber&
   s_t &= \tanh(c_t)\circ o
\end{align}

\noindent where the input $i$, forget $f$, and output $o$ represent gates that are squashed by the sigmoid into vectors of values between 0 and 1. Multiplying the vectors determines how much of the other vectors to let into the current input state. $g$ is a candidate hidden state that is computed based on the current input and the previous hidden state. $c_t$ is used as the internal memory, which is a combination of the previous memory $c_{t-1}$ multiplied by the input gate, and the hidden state $s_t$ is a combination of the internal memory and the output gate.

The RNNs, in their original form, process the sequences in just one direction. The problem is that they may have difficulty extracting information from the future that may be important in the present. 
A BRNN network usually consists of two RNNs, in any variant (naive RNN, LSTM, or GRU\cite{cho2014learning}), each of which processes the sequence in one direction, capturing patterns that can be difficult to see by a unidirectional RNN. The outputs of the two RNN are combined either by concatenation, multiplication, average, or sum.

\section{Question Answering Architecture}
Knowledge bases are a rich source of information that is accessible by experts of formal query languages.
The potential of exploiting knowledge bases can be increased by allowing any user to query the ontology by posing questions in natural language.

In this paper, this problem is seen as the following Natural Language Processing task:
Given an RDF knowledge base $O$ and a question $Q_{nat}$ in natural language (to be answered using $O$), translate $Q$ into a SPARQL query $S_{Q_{nat}}$ such that the answer to $Q_{nat}$ is obtained by running $S_{Q_{nat}}$ on $O$.

The starting point is a training set containing a number of pairs $\langle Q_{nat}, G_{Q_{nat}} \rangle$, where $Q_{nat}$ is a natural language question, and $G_{Q_{nat}}$ is a SPARQL query, called the \textit{gold query}.
The gold query is a SPARQL query that models (i.e., allows to retrieve from $O$) the answers to $Q_{nat}$.
The training set has to be used to learn how to answer questions posed in natural language using  $O$, so that, given a question in natural language $Q_{nat}$, the QA system can generate a query $S'_{Q_{nat}}$ that is equivalent to the gold query $G_{Q_{nat}}$ for $Q_{nat}$, i.e., such that $answers(S'_{Q_{nat}}) = answers(G_{Q_{nat}})$.
Basically, we compare the answers, and we are not interested in reproducing syntactically the gold query.
We approach this problem as a machine translation task, that is, we compute $S'_{Q_{nat}}$ as $S'_{Q_{nat}} = Translate (Q_{nat})$, where $Translate$ is the translation implemented by our QA system, called \oursystem.

Most of the solutions proposed up to now to convert from natural language to SPARQL make use of various techniques, either using patterns or deep neural networks (see Section \ref{sec:related}).
For obvious reasons (e.g., large size of the knowledge bases, frequent updates), all datasets comprise only a part of the input vocabulary, generating  the problem of \emph{Words Out Of Vocabulary} (WOOV) of the training set. Systems affected by the WOOV problem have difficulty dealing with words not seen during the training phase because they do not know how to map those words to the output vocabulary. For example, let's assume we have a training set containing the \emph{"Abraham Lincoln"} words and a system trained on it. If we want to translate the question \emph{When Abraham Lincoln was born?}; the system will be able to identify the right KB resource (dbr:Abraham\_Lincoln for DBpedia), but on the other hand, the system will fail to translate a question using the same pattern, but changing \emph{"Abraham Lincoln"} by something not present in the vocabulary, let say \emph{"Barack Obama"}.

\begin{figure}[t!]
  \caption{Architecture to translate questions into SPARQL.}
  \centering
    \includegraphics[scale=0.52]{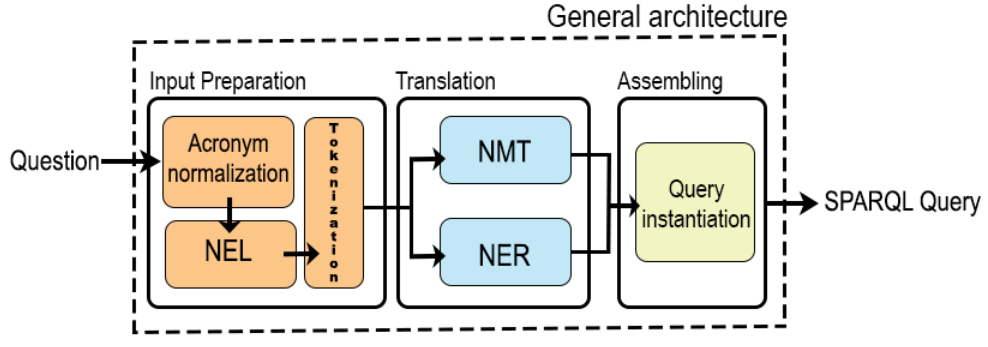} \vspace*{-0.1cm}
 \label{fig:arch} \vspace*{-0.5cm}
\end{figure}

To reduce the impact of the WOOV and improve the training time of the entire process, we introduce in \oursystem some remedies, including  a new format to represent an NL to SPARQL datasets. 
In particular, \oursystem implements a neural-network-based architecture (see Figure~\ref{fig:arch}) for question answering that accomplishes the objective by resorting to a novel combination of tools.
The architecture is composed of three main modules: \textit{Input preparation, Translation, and Assembling}, as shown in Figure \ref{fig:arch}. Each module is described in the following.

\subsection{Input preparation}
In this phase, the input sentence is processed in such a way that it is polished to attenuate linguistic noise (e.g., shifts in spelling, grammar, punctuation, entity identification) and also recast to be used as input for the subsequent phase.

\subsubsection{Acronyms normalization.}
In this step, acronyms are converted to the corresponding names.
For example, the $UK$ acronym becomes  \textit{United Kingdom}.
Acronyms regularly refer to KB resources inside the SPARQL query, and we need to replace them with the full name, which is more similar to the resource identifier in the KB.
In our implementation of the approach, this is particularly useful for handling acronyms of countries. To perform this task, we rely on two libraries. The first one is $spaCy$ \cite{spacy}, a well-known tool for NLP tasks. This library helps us to identify the acronyms thanks to their powerful NER mechanism. Then we use the \textit{Country Converter} (COCO) \cite{Stadler2017} library to obtain the original name.

\subsubsection{Fixing entities with NEL.}
The objective of applying this preprocessing step is to identify the entities in $Q_{nat}$ and replace them with the label used in the KB (DBpedia in our case) because our approach heavily relies on the correct spelling of the entities. 
For example, in the question: \textit{Where was the president Kennedy born?}, we look to transform the entity \textit{"Kennedy"} to \textit{"John F. Kennedy"} which is the label used by DBpedia to identify the resource. In our implementation, we face this problem by using \textit{Named Entity Linking} (NEL), also referred to as \textit{Named Entity Disambiguation}, which is the task of linking entity mentions in text with their corresponding entities in a target KB \cite{shen2014entity}. To this end, we employed \textit{DBpedia Spotlight} \cite{isem2013daiber}, a tool for automatically annotating mentions of DBpedia resources in text. Spotlight is available through an online $API$ that receives the text to be annotated and returns the annotation information together to a similarity score of each entity. In our case, we chose to process the resources with a similarity greater or equal to 0.8.

\subsubsection{Tokenization.}
The tokenization process is a fundamental step in almost all NLP methods. It is the task of chopping a text up into pieces called $tokens$, which are usually words. During this process, $Q_{nat}$ is cleaned by filtering out undesired characters such as punctuation marks and converted into a sequence of words $S$. We used the Keras preprocessing library. $S$ is converted then into a sequence of word embeddings. To mitigate the dependency on the input vocabulary (English) and reduce the impact of the OOV words \cite{goldberg2017neural}, we use the pre-trained word embeddings of $FastText$ \cite{bojanowski2017enriching}. $FastText$ is a library providing token embeddings also for words out of the input vocabulary.

\subsection{Translation}  \label{par:translation}
In this stage, the pre-processed $Q_{nat}$ is analyzed to recognize both its structure and the named entities that serve to build $S’_{Q_{nat}}$ in the subsequent assembling phase.
To do such recognition, we employ two neural networks performing two key NLP tasks:
$(i)$ a Neural Machine Translation task (NMT), and $(ii)$ a Named Entity Recognition (NER) task.

In the NMT task, $Q_{nat}$ is translated in a \textit{query template}.
A query template is the skeleton of a SPARQL query where some of the KB resources are replaced by placeholders.  
In general, in an NMT task, given a source $X = (x_1, x_2, ..., x_n)$ sequence and a target $Y = (y_1, y_2, ..., y_m)$ sequence, the aim is to model the conditional probability of target words given the source sequence \cite{bahdanau2014neural}. 
In our approach, an NMT neural network takes as input the sequence generated in the \textit{Preprocessing} phase and translates it into a SPARQL query template  
(details on the template's format are given later).

In the NER task, the entities present in $Q_{nat}$ are identified and classified using a dedicated neural network.
In a NER task (also known as \textit{Entity Extraction}), named entities present in a text are associated with predefined categories, such as \emph{individuals, companies, places, etc. }
This additional semantic knowledge helps to understand the role of words in a given text \cite{grishman1996message}.
As in most of the literature, in our implementation, we also adopt the \emph{BIO} notation \cite{ramshaw1999text} to tag a text, which differentiates the beginning $``B"$ and the interior $``I"$ of the entities while $``O"$ is used for non-entity tokens. 
Our NER network takes as input the sequence of tokens generated by the $Preprocessing$ phase and returns a tagging sequence in BIO notation, where the named entities correspond to possible KB resources within a SPARQL query.
The outputs of NMT and NER will then be combined in the assembling phase to build $S’_{Q_{nat}}$.

\subsubsection{Training set format.} \label{par:tformat}
The NMT and NER networks are trained together using the same input, obtained by converting the original training set in a novel format called $QQT$ format.
This pre-processing step has a two-fold objective. On the one hand, it aligns the inputs of both NMT and NER tasks; and, on the other hand, it reduces the size of the output vocabulary and helps to mitigate the impact of the OOV words during the translation.
Translating entities to URIs can be hard to learn from mere examples. A system can fail if the NER task is not simple and there are a lot of words that are Out Of the Vocabulary (OOV) of the training set. OOV is a serious problem while querying ontologies.

A dataset in QQT is composed of a set of triples in the form $\langle Question,$ $QueryTemplate, Tagging\rangle$, where $Question$ is a natural language question, \emph{Tagging} marks which parts of $Question$ are entities, and $QueryTemplate$  is a SPARQL query template modified as follows: $(i)$ The KB resources are replaced by one or more variables; $(ii)$ A new triple is added for each variable in the form \textit{"?var rdfs:label placeholder"}. $Placeholders$ are meant to be replaced by substrings of $Question$ depending on $Tagging$.

\begin{table}[t]\small
\caption{ $\langle Q_{nat}, Q_{sparql} \rangle$ for \emph{Who painted the Mona Lisa?}}
\centering
\begin{tabular}{|p{4.5cm}|p{7.2cm}|}
\hline
\textbf{Question} & \textbf{Query}\\
\hline
Who painted the Mona Lisa? & select ?a where \{  dbr:Mona\_Lisa  dbo:author ?a \} \\
\hline
\end{tabular} \label{tab:qq} \vspace*{-0.4cm}
\end{table}

\begin{table}[b!]\small
 \vspace*{-0.4cm}
\caption{QQT triple for \emph{Who painted the Mona Lisa?}}
\centering \vspace*{-0.2cm}
\begin{tabular}{|p{4.4cm}|p{5.4cm}|p{2cm}|}
\hline
\textbf{Question} & \textbf{QueryTemplate} & \textbf{Tagging} \\
\hline
Who painted the  Mona Lisa? & select ?a where \{  ?w dbo:author ?a .  ?w rdfs:label \$1 \} & O O O B I O \\
\hline
\end{tabular}
\label{tab:qqt}
\end{table}

In Table \ref{tab:qq}, we show an example of a $\langle Q_{nat}, Q_{sparql} \rangle$ pair for the question \textit{Who painted the Mona Lisa?}, while Table \ref{tab:qqt} shows the corresponding $\langle Question,$ $QueryTemplate, Tagging\rangle$ triple in the QQT format. 
In Table \ref{tab:qqt}, the term $\$1$ denotes a placeholder where the number index 1 means that $\$1$ has to be replaced by the first entity occurring in the question, \textit{Mona Lisa} in this case, as represented by BIO tagging notation. 
Note that in the QQT format, the query template does not contain any KB resource, so the learning model does not need to understand that \textit{Mona Lisa} stands for \textit{dbr:Mona\_Lisa}. With this representation, the output vocabulary of the NMT model is reduced, and the network is more tolerant to the WOOV problem.

\subsubsection{The networks.}
As we have seen, the architecture has two important parts that are NMT and NER. To develop the NMT neural network, we decided to use the standard \textit{Encoder-Decoder} approach, with BiLSTM and Luong Attention \cite{luong2015effective}, which has shown high performance in the literature. 
The Encoder extracts semantic information from the question and encodes it into a fixed-length vector $V$, and then the Decoder tries to decode $V$ into a sequence in the output language. 
On the other hand, to develop the NER network, we used the BiLSTM-CRF approach proposed by \cite{huang2015bidirectional}, which assigns a tag (BIO notation) to each token in the input sequence. 
The two models share the fact that they have a BiLSTM-based encoder to obtain the semantic information from the input sequence. For this reason, we decided to do a joint training of the two models looking to improve the training times and make the two networks help each other. 

\begin{figure}[t!]
\caption{Network for joint training of NMT and NER.} 
  \centering
    \includegraphics[scale=0.62]{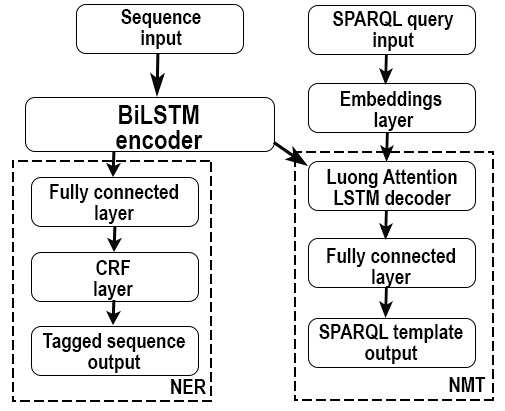}
  \label{fig:netw}
 \vspace*{-0.2cm}
\end{figure}

The proposed approach, depicted in Figure~\ref{fig:netw}, has a single encoder composed of one BiLSTM layer with two branches connected to it. The first branch uses a CRF layer responsible for determining $ p(l | x) $, which refers to the probability of calculating a tagging sequence $l$ given an input sequence $x$. The second one is an NMT decoder, composed of one LSTM layer and the attention mechanism followed by a fully connected network calculating the probability $p(y|x)$ that the sequence $y$ correctly translates $x$ in $QueryTemplate$.

\subsection{Assembling}
The last step of the proposed architecture is the creation of $Q’_{sparql}$. Here the placeholders in $Q’_{temp}$ are replaced by the corresponding entities.
For example, let's assume we have the pair:
$\langle$\textit{ Show me all Italian movies, SELECT DISTINCT ?uri WHERE \{  ?uri to dbo:Film; dbo:country dbr:Italy \} } $\rangle$ in the form $\langle Q_{nat}, Q_{sparql} \rangle$. Our system will translate $Q_{nat}$ to the following query template:

{\small
\begin{verbatim}
SELECT DISTINCT ?uri WHERE { ?uri a dbo:Film; dbo:country ?v 
. ?v rdfs:label $1 }
\end{verbatim}} \vspace*{-0cm}
\noindent and the NER task will identify that $ Italian $ is the named entity. So the query instantiation step will produce:
{\small
\begin{verbatim}
SELECT DISTINCT ?uri WHERE { ?uri a dbo:Film; dbo:country ?v 
. ?v rdfs:label ”Italy”@en }
\end{verbatim}
}
\noindent which is a SPARQL query equivalent to $Q_{sparql}$.

\section{Experiments}

\paragraph{Experiment Setup.} We have implemented our models using Keras, a well-known framework for machine learning built on top of TensorFlow. %
We trained the networks using Google Collaboratory, a virtual machine environment hosted in the cloud and based on Jupyter Notebooks. With Google Collaboratory, it is possible to execute the code using a GPU configuration, speeding up the training task. 
The environment provides 12GB of RAM and connects to Google Drive. 
We considered the well-known and publicly available datasets for QA over DBpedia ontology: Monument, QALD-9, and LC-QuAD v1.
To assess the systems, we adopted the macro precision, recall, F1-score, and F1 QALD measures as proposed in the QALD-9 challenge.

\begin{table}[b!]\small
\caption{Comparison on Monument.} 
\centering
\begin{tabular}{cllllll} 
\hline
\textbf{}                        & \multicolumn{3}{c}{\textbf{Mon300}}                                             & \multicolumn{3}{c}{\textbf{Mon600}}                                              \\ 
\cline{2-7}
                                 & \multicolumn{1}{c}{\textbf{P}} & \multicolumn{1}{c}{\textbf{R}} & \multicolumn{1}{c}{\textbf{F1}} & \multicolumn{1}{c}{\textbf{P}} & \multicolumn{1}{c}{\textbf{R}} & \multicolumn{1}{c}{\textbf{F1}}  \\ 
\hline
\multicolumn{1}{l}{NSpM}         & 0.860                   & 0.861                   &  0.852                     & 0.929                   & 0.945                   & 0.932                     \\ 
\hline
\multicolumn{1}{l}{\textbf{\oursystem}} & 0.783                     & 0.791                     & 0.786                      & 0.780                     & 0.781                     & 0.780                       \\
\hline
\end{tabular}
\label{tab:exp1}
 \vspace*{-0.3cm}
\end{table}

\subsection{Evaluation on Monument dataset}
The Monument dataset was proposed as part of the Neural SPARQL Machines (NSpM) \cite{soru2017sparql} research. It contains 14,778 question-query pairs about the instances of type Monument present in DBpedia.
For the sake of comparison with the state-of-the-art, we have trained the Learner Module of NSpM as it was done in \cite{soru2017sparql}, where the authors proposed two instances of the Monument dataset that we will denote by Monumet300 and Monument600 containing 8,544 and 14,788 pairs, respectively. In both cases, the dataset split fixes 100 pairs for both validation and test set, keeping the rest for the training set. All the data is publicly available in the NSpM GitHub project (https://github.com/LiberAI/NSpM/tree/master/ data).
To train our system, we first performed hyperparameter tuning focused on three metrics: embedding size of the target language, batch size, and LSTM hidden units. 
We performed the tuning using a grid search method. The number of epochs was set to 5, shuffling the dataset at the end of each one. After tuning, we set the hyperparameters as follows: embedding size is set to 300, LSTM hidden units are set to 96, and batch size is set to 64.
From the results in Table \ref{tab:exp1}, we see that our system performs reasonably well, F1-score greater than 0.78, and NSpM has better results. 

\begin{table}[b]
 \vspace*{-0.3cm}
\small
\caption{Comparison on OOV entities dataset.} 
\centering
\begin{tabular}{cllllll} 
\hline
\textbf{}                        & \multicolumn{3}{c}{\textbf{Mon300 model}}                                             & \multicolumn{3}{c}{\textbf{Mon600 model}}                                              \\ 
\cline{2-7}
                                 & \multicolumn{1}{c}{\textbf{P}} & \multicolumn{1}{c}{\textbf{R}} & \multicolumn{1}{c}{\textbf{F1}} & \multicolumn{1}{c}{\textbf{P}} & \multicolumn{1}{c}{\textbf{R}} & \multicolumn{1}{c}{\textbf{F1}}  \\ 
\hline
\multicolumn{1}{l}{NSpM}         & 0.097                   & 0.123                   & 0.101                    & 0.110                   & 0.110                   & 0.110                     \\ 
\hline
\multicolumn{1}{l}{\textbf{\oursystem}} & 0.831                     & 0.8291                     & 0.830                      & 0.863                     & 0.879                     & 0.866                       \\
\hline
\end{tabular}
\label{tab:exp2}
\end{table}

We have investigated the cases in which our system could not provide an optimal answer, and we discovered that the performance of our approach is mainly affected by problems in the dataset. We found a set of questions that lacks context to determine specific expected URIs. For example, for the question ``What is Washington Monument related to?'' our system uses  ``Washington Monument'', but the gold query uses the specific URI: \textit{Washington\_Monument\_(Baltimore)}. Note that there is no reference to Baltimore in the question text, and there are Washington Monuments also in Milwaukee and Philadelphia, according to DBPedia. 
However, NSpM often uses the specific URI of the gold query.
Thus, we decided to devise a tougher experiment to better understand the issue. 
We used the templates of NSpM and a randomly selected set of unseen monument entities extracted from DBpedia to create a new test set of 200 pairs. 
Results in Table \ref{tab:exp2} show that our approach confirms the same good performance (F1 score greater than 0.80), demonstrates a better generalizing power being basically resilient to the presence of unseen entities and also performs better than NSpM.

\subsection{Evaluation on QALD-9}
The \textit{Question Answering over Linked Data} (QALD) is a series of challenges that aim to provide benchmarks for assessing and comparing QA systems on DBpedia~\cite{ngomo20189th}. 
We considered the benchmark proposed as part of the ninth edition of QALD, known as QALD-9 (https://github.com/ag-sc/QALD/tree/master/9/ data).
The dataset contains 558 question-query pairs in 11 different languages. The data is split into 408 training and 150 testing questions, and we focus on the ones expressed in English.
It is important to note that this dataset is very challenging to be approached using learning techniques, given the very small training set not covering all questions types of the test set.

We reproduced the original QALD-9 setting to compare our system with the systems that participated in the competition. To train our network, we performed a cross-validation process to adjust the model parameters, motivated by the small number of examples in the training set. We use the same settings as with the Monument dataset and set the epochs to 30. 
Table \ref{tab:exp3} shows the performance of the QALD-9 challengers and the NSQA~\cite{kapanipathi2020question} system proposed by IBM, here our model was able to learn from small data, ranking virtually in the first position.

\begin{table}[t]
\small
\caption{Comparison on QALD-9.} 
\centering
\begin{tabular}{lcccc} 
\hline
                      & \textbf{P} & \textbf{R} & \textbf{F1} & \textbf{F1 QALD}  \\ 
\hline
Elon         & 0.049      & 0.053      & 0.050       & 0.100             \\
QASystem     & 0.097      & 0.116      & 0.098       & 0.200             \\
TeBaQA       & 0.129      & 0.134      & 0.130       & 0.222             \\
wdaqua-core1 & 0.261      & 0.267      & 0.250       & 0.289             \\
gAnswer      & 0.293      & 0.327      & 0.298       & 0.430    \\
NSQA      & 0.314      & 0.321      & 0.308       & 0.453    \\
\textbf{\oursystem}          & 0.310      & 0.3248      & 0.306       & \textbf{0.466 }             \\
\hline
\end{tabular}
\label{tab:exp3}
 \vspace*{-0.3cm}
\end{table}

Given the very small set of questions, we decided to expand the training set to further improve our system and better understand its behavior. Thus, we created templates from the gold questions by annotating all the named entities with spaCy and checking them manually; then, we generated new questions by replacing the annotated entities with others randomly selected up to creating a total of 1816 pairs. 
The obtained set is the \textit{expanded} training set. 
Further, we applied the same query generation process to the pairs of the test set and added the new pairs to the expanded training set. No pair or gold query from the original test set was added. In this way, we created a new benchmark (labeled \textit{expanded w/test}) containing 2331 examples. 
As we can see in Table \ref{tab:exp4}, expanding the dataset with the same question types (\emph{expanded dataset}) does not improve the results, rather it could be harmful because there is more repeatability in a training set that is not representative of the test set, conducting to less generalization.
Moreover, when the training set is expanded with patterns from the original test set (see the expanded w/test row), our system reaches a great performance (F1 QALD of 0,79). 

\begin{table}[b]
\small
\caption{\oursystem trained on QALD-9 expanded.} 
\centering
\begin{tabular}{lcccc} 
\hline
                      & \textbf{P} & \textbf{R} & \textbf{F1} & \textbf{F1 QALD}  \\ 
\hline
\textbf{expanded}         & 0.316      & 0.321      & 0.309       &0.460             \\
\textbf{expanded w/test}     & 0.687      & 0.688      & 0.682       & 0.794             \\
\hline
\end{tabular}
\label{tab:exp4}
\end{table}

\begin{table}[t]\small
 \vspace*{-0.3cm}
\caption{Comparison on LC-QuAD v1.}
\centering
\begin{tabular}{lccc} 
\hline
                 & \textbf{P}     & \textbf{R}     & \textbf{F1}     \\ 
\hline
WDAqua & 0.220          & 0.380          & 0.280           \\
QAMP    & 0.250          & \textbf{0.500} & 0.330           \\
NSQA    & 0.448          & 0.458          & 0.444           \\
\textbf{\oursystem}     & \textbf{0.495} & 0.492          & \textbf{0.491}  \\
\hline
\end{tabular}
\label{tab:lcquad}
\end{table}

\begin{table}[b]\small
\caption{Ablation: F1 scores Monument, Monument OOV.} 
\centering
\begin{tabular}{lcc|cc}
\hline
                          & \multicolumn{2}{l|}{\textbf{Mon300 model}} & \multicolumn{2}{c}{\textbf{Mon600 model}} \\ \cline{2-5} 
\textbf{}                 & \textbf{test set}  & \textbf{OOV} & \textbf{test set} & \textbf{OOV} \\ \hline
\textbf{\oursystem} & 0.7862             & 0.8297       & 0.7804            & 0.8656       \\
\textbf{w/o Acro}                  & 0.7762             & 0.8297       & 0.7804            & 0.8556       \\
\textbf{w/o NEL}                  & 0.8140             & 0.7950       & 0.7912            & 0.8023       \\ \hline
\end{tabular}
\label{tab:abl1}
\end{table}

\subsection{Evaluation on LC-QuAD v1}

Large-Scale Complex Question Answering Dataset (LC-QuAD v1)~\cite{trivedi2017lc} (https:// github.com/AskNowQA/LC-QuAD) was created in 2017 to provide a sufficiently large, complex, and varied dataset for the application and evaluation of machine learning-based QA approaches. The benchmark has 5000 question-query pairs, where the train set has 4000 questions, and the test set the remaining 1000 pairs. LC-QuAD v1 is based on DBpedia, and more than 80 \% of the questions contain two or more relationships increasing complexity.

To evaluate our system in LC-QuAD v1, we use the same procedure as with QALD-9 to adjust the model parameters.
Our system outperforms compared methods, see Table \ref{tab:lcquad}.

\subsection{Ablation study}
For completeness, we report the result of an ablation study measuring the impact of the various modules of our system. 
We removed one module at a time and recalculated the metrics for all the considered datasets. 
For QALD-9, we used both the original and the \emph{expanded w/test} datasets for training, so to provide a more extensive analysis.

\subsubsection{Acronym normalization.}
As shown in Table \ref{tab:abl1}, the lack of the acronyms normalization module basically does not affect the final results for the two Monument dataset variants. 
This low impact is due to the lack of acronyms in the monument test sets. 
The phenomenon also applies to the LC-QuAD v1 dataset, where after removing the analyzed module, the results are the same.
However, in the case of QALD-9, shown in Table \ref{tab:abl2}, we can see that the F1 QALD measure for the two models without acronym normalization is lower than the previous values. 
This behavior is due to the presence of acronyms that the system without the module under assessment was not able to deal with. 
All in all, acronyms normalization improves the overall performance of the system in case acronyms are part of the input.
It is important to note that the NEL task can sometimes adjust acronyms.

\subsubsection{Named Entity Linking.}
By looking at Table \ref{tab:abl1}, we note that the lack of entity linking has little impact on the system results in Mon600 and Mon300 original test sets. 
This is because the entities found in the questions almost always coincide with the text value of the \emph{rdfs:label} KB predicate used by our approach to compose the queries, and the NEL module can make mistakes. 
For the Monument OOV, the situation changes, and we can appreciate a clear drop in the F1 score values when there is no entity linking. 
On the other hand, after removing the NEL module, the changes in the LC-QuAD v1 scores are really small. Like in the Monument dataset, the entities are well represented in the questions, and this applies strongly to those questions that \oursystem can answer well.
Concerning the more involved QALD-9 dataset, we report a drop in the F1 QALD measure without NEL (see Table~\ref{tab:abl2}).
The performance decreases by about 3.4 \% on the expanded dataset (right column). 
This shows the usefulness of NEL on more involved questions.

\subsubsection{Joint training.}
Joint training of NER and NMT networks is a novelty of our approach. 
To validate it, we trained each network separately using all the datasets considered in this paper, and then we compared it with the system using joint training. 
The first benefit of the joint training is to save execution time since we do not have to wait for the NMT network training to start with the NER network and vice versa. 
In the experiment, we obtained a training time reduction of 40 \% by applying joint training.
The second and expected benefit is to obtain a better generalization from the data since the networks for NER and NMT might help each other.
In the Monument dataset, the system performance does not change significantly with joint training. The impact is more evident with the more involved dataset QALD-9. 
By training the networks separately, we were unable to obtain an F1 QALD value greater than 0.390, which is much less than the 0.466 obtained with the joint training. 
For recall, precision, and F1-score, we obtained values of 0.264, 0.260, and 0.255, respectively, which are also smaller than our best results with joint training. This analysis confirms the positive impact of the joint training technique.

\begin{table}[t]\small
\caption{F1 QALD scores for the ablation study on QALD-9.}
\centering
\begin{tabular}{lcc}
\hline
\textbf{}                 & \textbf{QALD-9 test set} & \textbf{Expanded w/test} \\ \hline
\textbf{\oursystem} & 0.4669         & 0.7948          \\
\textbf{w/o Acro}                  & 0.4579         & 0.7810           \\
\textbf{w/o NEL}                   & 0.4567         & 0.7674          \\ \hline
\end{tabular}
\label{tab:abl2}
\vspace*{-0.2cm}
\end{table}

\section{Related Work}\label{sec:related}

\paragraph{Pattern-based.}
The idea of employing query patterns for mapping questions to SPARQL-queries was already exploited in the literature~\cite{pradel2013natural,steinmetz2019natural}. 
The approach presented by Pradel and Ollivier~\cite{pradel2013natural} also adopts named entity recognition but applies a set of predefined rules to obtain all the query elements and their relationships. 
The approach by Steinmetz et al.~\cite{steinmetz2019natural} has 4 phases, firstly, the question is parsed, and the main focus is extracted, then general queries are generated from the phrases in natural language according to predefined patterns, and finally, make a subject-predicate-object mapping of the general question to triples in RDF. 
Despite both of the above-mentioned approaches performed well in selected benchmarks, they rely on patterns and rules defined manually for all existing types of questions. 
A limit that is not present in our proposal.

\paragraph{Deep Learning-based.}
The \emph{Neural SPARQL Machines (NSpM)}~\cite{soru2017sparql} approach is based on the idea of modifying the SPARQL queries to treat them as a foreign language. They encoded the brackets, URIs, operators, and other symbols, making the tokenization process easier. The resulting dataset was introduced in a Seq2Seq model responsible for performing the question-query mapping. 
The same authors created the \emph{DBNQA dataset} \cite{hartmannMarxSoru2018}, and their model was tested on a subdomain referring to monuments and evaluated using the purely syntactic BLEU score~\cite{soru2017sparql}. As a consequence, it performs well in reproducing the syntax of the gold query but is less able to generalize to unseen questions and OOV words wrt. our approach. 
 
The query building approach by Chen et al.~\cite{chen2020formal} features two stages. The first stage consists of predicting the query structure of the question and leverages the structure to constrain the generation of the candidate queries. The second stage performs a candidate query rank. As in our approach, Chen et al. use BiLSTM networks, but query representation is based on abstract query graphs. 
They evaluate the candidate queries at string level w.r.t to the gold query, ignoring the evaluation of selecting target variables as they do not execute the queries. 
The usage Encoder-Decoder model based in LSTM with an attention mechanism to associate a vocabulary mapping between natural language and SPARQL was also proposed in the literature by~\cite{luz2018semantic}, obtaining good results. 
As evidenced in the paper, they do nothing to deal with OOV words, so it is a weakness against our approach. 

Kapanipathi et al.~\cite{kapanipathi2020question} proposed a system called NSQA that performs several steps to obtain the final query. First, the question is semantically parsed to an Abstract Meaning Representation (AMR). Next, they align the AMR with the KB by applying entity linking followed by a path-based algorithm to generate triples that have a one-to-one correspondence to the triples in the final SPARQL query. Finally, the triples are converted to a first-order logic representation to be used by a Logical Neural Network responsible to performs the reasoning to generate the SPARQL query. This system does not require end-to-end training data and has been demonstrated to work well on QALD-9 and LC-QuAD v1 datasets. We wanted to compare NSQA with our system using the Monument dataset, but it is not an open-source project. Then, we asked the authors for help to obtain the necessary data for performing the comparison, but finally, we did not receive the information.
Also, we report that eight different models based on RNNs and CNNs were compared by Yin, Gromann, and Rudolph~\cite{yin2019neural}, and ConvS2S \cite{gehring2017convolutional} proved to be the best.

For completeness, we studied another related line of works that aims to translate the natural language questions into SQL queries and BASH commands. 
In the Seq2SQL approach~\cite{zhong2017seq2sql}, an LSTM Seq2Seq model is used to translate from natural language to SQL queries. The interesting thing about this approach is that they use \emph{Reinforcement Learning} to guide the learning. 
Yu et al.~\cite{yu2018spider} introduce a large-scale, complex, and cross-domain semantic parsing and text-to-SQL dataset to train different models to convert text to SQL queries. Most of the models were based on a Seq2Seq architecture with attention, demonstrating an adequate performance. Another interesting approach for text-to-SQL generation was introduced by Zhang et al.~\cite{zhang2019editing}. They implement a Seq2Seq model with Luong's attention, using BiLSTMs and BERT embeddings. The approach performs well on SParC and Spider datasets, outperforming the related work in some cases. 

Lin et al.~\cite{lin2018nl2bash} address the topic of natural language to BASH commands translation. They propose NL2Bash, a new dataset in the ambitious domain of controlling the operating system using natural language, containing more than 9000 pairs.
To demonstrate that NL2Bash is challenging, and establish a baseline for the proposed dataset, the authors applied and evaluated three approaches, $(i)$ Seq2Seq model, $(ii)$ CopyNet, and $(iii)$ Tellina, demonstrating decent performances. Also, this approach uses a Seq2Seq model, but it is unclear how the OOV words are considered.

Our architecture addresses many of the issues connected with the translation (e.g., acronyms, entity linking), resorting to specific tools, an aspect that is not present in mentioned works. Moreover, existing approaches based on NMT do nothing special to deal with out-of-vocabulary words.

\section{Conclusions and Future Work}
The paper presents a novel QA system based on deep neural networks to query knowledge bases by using natural language. 
The system roots on effective ideas presented in the literature but addresses an issue that can hinder the application of current state-of-the-art QA implementations in real-world scenarios: the inability to deal with the OOV words.
To this end, our system resorts to several well-known NLP tools and combines them in an original way: translation and NER are co-trained in the RNN architecture, and a novel format for the training set is introduced.
Our system showed effective results on the Monument, QALD-9, and LC-QuAD v1 publicly available datasets and demonstrated a more general and robust behavior on unseen questions among the compared system. 
In future work, we plan to improve translation performance by considering other NLP tools, such as Transformers and BERT embeddings.

%
%
%
\bibliographystyle{splncs04}
\bibliography{corr}

\end{document}